\documentclass{article}

\usepackage{arxiv}

\usepackage[utf8]{inputenc} 
\usepackage[T1]{fontenc}    
\usepackage{hyperref}       
\usepackage{url}            
\usepackage{booktabs}       
\usepackage{amsfonts}       
\usepackage{nicefrac}       
\usepackage{microtype}      
\usepackage{lipsum}
\usepackage{graphicx}

\usepackage{graphicx}
\usepackage[T1]{fontenc}
\usepackage{array}
\usepackage{float}
\usepackage{subfig}
\usepackage{booktabs}
\usepackage{multirow}
\usepackage{amsmath}
\usepackage{amsthm}
\usepackage{epsfig}
\usepackage{amssymb}
\usepackage{stackrel}
\usepackage{nomencl}
\usepackage{breqn}
\usepackage{colortbl}
\usepackage[table]{xcolor}

\graphicspath{ {./images/} }

\title{Enhancing reinforcement learning by a finite reward response filter with a case study in intelligent structural control}

\author{
 Hamid Radmard Rahmani\\
 (Corresponding author)\\
  Institute of Structural Mechanics \\ 
  Bauhaus-Universit\"at Weimar, Germany\\
  \texttt{radmard.rahmani@gmail.com} \\
   \And
 Carsten Koenke \\
  Institute of Structural Mechanics \\ 
  Bauhaus-Universit\"at Weimar, Germany\\
  \And
 Marco A. Wiering \\
  School of Coumputing and Information\\
   University of Groningen,  Netherlands\\
}

\begin{document}
\maketitle
\begin{abstract}
In many reinforcement learning (RL) problems, it takes some time until a taken action by the agent reaches its maximum effect on the environment and consequently the  agent receives the reward corresponding to that action by a delay called \emph{action-effect delay}. Such delays  reduce the performance of the learning algorithm and increase the computational costs, as the reinforcement learning agent values the immediate rewards more than the future reward that is more related to the taken action. This paper addresses this issue by introducing an applicable enhanced Q-learning method in which  at the beginning of the learning phase, the agent takes
a single action and builds a function that reflects the environment's response to that action, called the \emph{reflexive $\gamma$ - function}. During the training phase, the agent utilizes the created \emph{reflexive $\gamma$ - function} to update the Q-values.
We have applied the developed method to a structural control problem in which the goal of the agent is to reduce the vibrations of a building subjected to earthquake excitations with a specified delay. Seismic control problems are considered as a complex task in structural engineering because of the stochastic and unpredictable nature of earthquakes and the complex behavior of the structure. Three scenarios are presented to study the effects of zero, medium, and long action-effect delays and the performance of the Enhanced method is compared to the standard Q-learning method. Both RL methods use a neural network to learn to estimate the state-action value function that is used to control the structure.
The results show that the enhanced method significantly outperforms the performance of the original method in all cases, and also improves the stability of the algorithm in dealing with action-effect delays.

\keywords{Reinforcement learning \and Neural networks \and Structural control \and Earthquake \and Seismic control\and Smart structures}
\end{abstract}

\section{Introduction}
Reinforcement learning (RL) \cite{RichardS.SuttonReinforcementLearningintroduction1998,WieringReinforcementLearningStateoftheArt2012} is an area of machine learning concerned with
an  \emph{agent} that learns to perform a task so that
it maximizes the (discounted) sum of future \emph{rewards}. For this,
the agent interacts with the  \emph{environment}
by letting it select actions based on its observations. After each
selected action, the environment changes and the agent receives a
reward that indicates the goodness of the selected action.
The difficulty is that the agent does not only want to obtain the
highest immediately obtained reward, but wants to optimize
the long-term reward intake (return). The return or discounted
sum of future rewards depends on all subsequent selected actions and visited states during the decision-making process.

In the literature, RL methods have been successfully applied to make
computers solve different complex tasks, which are usually done by  humans. Developing autonomous helicopters \cite{NgAutonomousInvertedHelicopter2006}, playing Atari games \cite{MnihHumanlevelcontroldeep2015}, controlling traffic network  signals \cite{Aslani:2018}, and mimicking human's behavior in robotic tasks \cite{KoberReinforcementlearningrobotics2013} are some examples for applications of RL methods in solving real-world problems.

As a development, we have applied RL to the seismic control problem in structural engineering as one of the most complex structural problems due to the stochastic nature of the earthquakes and their wide frequency content, which makes them unpredictable and also destructive to many different structures.

The past experience indicates that earthquakes are responsible for damaging many buildings and cause
many deaths each year. The worst earthquakes in history cost more
than 200,000 human lives.

Providing acceptable safety and stability
for buildings and other structures during earthquakes is a challenging
task in structural engineering. The engineers have to take into account
the stochastic nature of the earthquake excitations and many other
uncertainties in the loads, the materials, and the construction. The creation
of structural control systems is a modern answer to dealing with this problem.
A controlled structure is equipped
with control devices that will be triggered in case of occurring
natural hazards such as heavy storms or earthquakes. 
There are four types of structural control systems, namely passive,
active, semi-active, and hybrid systems. Passive systems do not
need a power supply to function. Base isolators and tuned mass dampers 
(TMDs)
are two examples of such systems. Active control systems need electricity
to apply forces to the structure, but the advantage is that the magnitude
and the direction of the force are adjustable. Semi-active systems
need to use less electricity compared to active controllers, which allows
the  power to be provided by batteries and other power storage types. Sometimes the optimal solution is achieved by combining
different control types to create a hybrid system. 
The advances in structural control systems have resulted in the birth
of a new generation of  structures called \emph{smart structures}. These structures
can sense their environment through sensors and use an intelligent
control system to stabilize the structure and provide
safety to the occupants.

The main part of the control system in a smart structure is the \emph{control
algorithm} that determines the behavior of the controller device during
the external excitations. This paper describes the use of reinforcement
learning (RL) \cite{RichardS.SuttonReinforcementLearningintroduction1998} combined with neural networks to learn to apply forces to
a structure in order to minimize an objective function. This objective
function consists of structural responses including the displacement, the velocity and the acceleration of some part of the structure over time.

The difficulty in intelligent structural control is that applied forces to stabilize the structure usually
reach to their maximum effect on the building after a specific time period. Moreover, the magnitude and the direction of the structural responses changes rapidly in the time during the earthquake excitations. These issues makes determination of the Q-values difficult for the agent and results in poor learning performance. To deal with this problem, we introduce a novel RL method that is based on a {\it finite reward response filter}. In our proposed method, the agent does not receive a reward value
only based on the change of the environment in a single time step, but
uses the combined rewards during a specific time period in the future
as feedback signal. We compare the proposed method to the conventional
way of using rewards for single time steps using the Q-learning algorithm \cite{WatkinsQlearning1992} combined with neural networks. For this,
we developed a simulation that realistically mimics a specific building
and earthquake excitations. The building is equipped with an active
controller that can apply forces to the building to improve its responses.
The results show that using the proposed finite reward response filter
technique leads to significant improvement in the performance of the controller compared to when it is trained by the conventional RL method.

The rest of this paper is organized as follows. Section 2 explains
the background of different control systems that have been used
for the structural control problem. Section 3 describes the used RL
algorithms and theories. Section 4 explains the novel method. 
Section 5 presents the case study by describing the structural model, dynamics of the environment, and the details of the simulation. Section 6 explains the experimental set-up. Section 7 presents the results, and finally
Section 8 concludes this paper.

\section{Background in Structural Control Systems}

In the literature, several types of control algorithms
have been studied for the structural control problem, which
can be categorized in four main categories: Classical, Optimal, Robust, and Intelligent. These different approaches will be shortly explained below.

{\bf Classical Control Systems}. Classical approaches are developed based on the  proportional-integral-derivative
(PID) control concept. Here, the controller continuously calculates
an error value as the difference between the desired set-point (SP) and
the measured process value (PV). The errors over time are used
to compute a correction control force
based on the proportional (P), integral (I), and derivative terms (D). 
Two types of structural controllers that use
the classical approach are feedback controllers and feedforward controllers
\cite{Dorf:2000:MCS:557022,AzelogluVibrationmitigationnonlinear2016,KaliannanAntcolonyoptimization2016,BalasDirectvelocityfeedback1979,ChaseRobuststaticoutput1999,IkedaActivemassdriver,FranklinFeedbackcontroldynamic1994}. 
Although such control systems may work well in particular cases,
they do not take the future effects into account when computing control
outputs, which can lead to sub-optimal decisions with negative 
consequences.

{\bf Optimal Control.} Optimal control algorithms aim to optimize an objective function while considering the problem constraints.
When the system dynamics are described by a set of linear differential
equations and the cost function is defined by a quadratic function,
the control problem is called a linear-quadratic control (LQC) problem \cite{KwakernaakLinearoptimalcontrol1972}.
Several studies have been carried out on optimal control of smart
structures. One of the main solutions for such control problems is
provided by the linear quadratic regulator (LQR) \cite{AdeliOptimalcontroladaptive1997}.
The LQR solves the control problem using a mathematical algorithm 
that minimizes the quadratic cost function given the problem
characteristics \cite{SontagMathematicalcontroltheory2013}. Such algorithms
are widely used for solving structural control problems. 
Ikeda et al. \cite{IkedaActivemassdriver} utilized an LQR control algorithm to
control the lateral displacement and torsional motions of a 10-story
building in Tokyo using two adaptive TMDs (ATMDs). 
This building had afterwards experienced
one earthquake and several heavy storms and showed a very
good performance in terms of reducing the motions. In another study, Deshmukh and Chandiramani
\cite{DeshmukhLQRControlWind2014} used an LQR algorithm to control
the wind-induced motions of a benchmark building equipped with a semi-active
controller with a variable stiffness TMD. 

Controllers often also have to deal with stochastic effects due to the mentioned uncertainties \cite{Daimultiwaveletneuralnetworkbased2015}.
Model predictive control (MPC) algorithms use a model to estimate
the future evolution of a dynamic process in order to find
the control signals that optimize the objective function. Mei et al.
\cite{MeiModelpredictivecontrol2002} used MPC together with
acceleration feedback for structural control under earthquake
excitations. In their study, they utilized the Kalman-Bucy filter to
estimate the states of systems and performed two case studies including a single-story and a three-story building, in which active tendon control devices were used. Koerber and King \cite{KoerberCombinedfeedbackfeedforward2013}
utilized MPC to control the vibrations in wind turbines.

Although these optimal control algorithms can perform very well,
they rely on having a model of the structure that is affected by the
earthquake. This makes such methods less flexible and difficult to use
for all kinds of different structural control problems.

{\bf Robust Control.} Robust control mainly deals with uncertainties and has the goal to
achieve robust performance and stability in the presence of bounded
modeling errors \cite{LevineControlsystemfundamentals2019}. In contrast
to adaptive control, in robust control the action-selection
policy is static \cite{AckermannRobusteRegelungAnalyse2013}.
$H_{2}$ and $H_{\infty}$ control are some typical examples of robust
controllers that synthesize the controllers to achieve stabilization
regarding the required performance. Initial studies on $H_{2}$
and $H_{\infty}$ control were conducted by Doyle et al. \cite{DoyleStatespacesolutionsstandard1989}.
Sliding mode control (SMC) is another common control method for
nonlinear systems in which the dynamics of the system can be altered
by a control signal that forces the system to ``slide'' along a
cross-section of the system's normal behavior. Moon et al. \cite{MoonSeokJ.SlidingModeControl2003}
utilized SMC and the LQC formulation for vibration control of a cable-stayed
bridge under seismic excitations. They evaluated the robustness of
the SMC-based semi-active control system using magnetorheological
(MR) dampers which are controlled using magnetic fields. 
In another research, Wang and Adeli \cite{WangAlgorithmschatteringreduction2012}
proposed a time-varying gain function in the SMC. They developed two
algorithms for reducing the sliding gain function for nonlinear
structures. 
Although robust control methods can be used to guarantee a robust
performance, they also require models of the structural problem
and often do not deliver optimal control signals for a specific situation.

{\bf Intelligent Control.}
Intelligent control algorithms  
are capable of dealing with complex problems consisting of a high degree of uncertainty using \emph{artificial intelligence} techniques.
The goal is to develop autonomous systems that can sense the environment and operate autonomously in unstructured
and uncertain environments \cite{PoznyakDifferentialneuralnetworks2001}.
Intelligent control uses various 
approaches such as fuzzy logic, machine learning, evolutionary computation, or combinations of these methods such
as neuro-fuzzy \cite{JangNeurofuzzysoftcomputinga1997} or genetic-fuzzy
\cite{JamshidiIntelligentcontrolsystems2001} controllers. 
In the last decade, algorithms such as neural
networks \cite{cabessa2014super,donnarumma2015programmer}, evolutionary
computing \cite{martinez2011evolutionary,cheng2015multicriteria},
and other machine learning methods \cite{you2014multiobjective,mesejo2015artificial}
have been used to develop intelligent controllers for smart structures.

An issue with fuzzy controllers is the difficulty to determine the best
fuzzy parameters to optimize the performance of the controller.
In this regard,
various methods have been developed to optimize the parameters
for fuzzy-logic control \cite{AhlawatMultiobjectiveoptimalstructural2001,AhlawatMultiobjectiveoptimalabsorber2003,AhlawatMultiobjectiveoptimalfuzzy2004,kim2006design}.
This optimization process can be based on
online and offline methods and has also been used for the vibration
control problem of smart structures \cite{AzelogluVibrationmitigationnonlinear2016,quiros2014use,SamaliPerformancefivestoreybenchmark2003,SoleymaniAdaptivefuzzycontroller2014}. 

Neural controllers have been investigated in a
few studies \cite{Bani-HaniKhaldoonNonlinearStructuralControl1998,Bani-HaniNeuralnetworksstructural1998,GhaboussiActivecontrolstructures1995}
in which neural networks are utilized to generate the control
commands. In these studies, the neural network has been trained
to generate the control commands based on a training set dictated
by another control policy. The disadvantage is that these methods
have little chance to outperform the original control algorithm, and
therefore can only make decisions faster.
Madan \cite{MadanVibrationcontrolbuilding2005} developed a 
method to train a neural controller to improve the responses
of the structure to earthquake excitations using a modified counter-propagation neural network. This algorithm learns to
compute the required control forces without a training set 
or model of the environment.

Rahmani et. al. \cite{Rahmaniframeworkbrainlearningbased2019} were the
first to use reinforcement
learning (RL) to develop a new generation of structural intelligent
controllers that can learn from experiencing earthquakes to optimize
a control policy. Their results show that the controller has a very good performance under different environmental and structural uncertainties.
In the next section, we will explain how we 
extended this previous research by using finite reward response filters
in the RL algorithm.

\section{Reinforcement Learning for Structural Control}
In RL,
the environment is typically formulated as a Markov decision process
(MDP) \cite{RichardS.SuttonReinforcementLearningintroduction1998}.
An MDP consists of a state space $S$, a set of actions $A$, a reward function $R(s,a)$ that maps state-action pairs to a scalar reward value, and a
transition function $T(s,a,s')$ that specifies the probability of
moving to each next possible state $s'$ given the selected action $a$
in the current state $s$. In order to make the intake of immediate
rewards more important compared to rewards received far in the future,
a discount factor $0 \leq \gamma \leq 1$ is used.

\subsection{Value functions and policies}

When the transition and reward functions are completely known, MDPs 
can be solved by using dynamic programming (DP) techniques \cite{Bellman:1957}. DP techniques rely on computing value functions
that indicate the expected return an agent will obtain given
its current state and selection action. The most commonly used
value function is the state-action value function, or Q-function.
The optimal Q-function $Q^*(s,a)$ is defined as:

\begin{equation}
Q^*(s,a) = \max_{\pi} E(r_{t}+\gamma r_{t+1}+\gamma^{2}r_{t+2}+\ldots+\gamma^{N} r_{t+N})
\end{equation}

Where $E$ is the expectancy operator which takes into account all 
stochastic outcomes, $r_t$ is the emitted reward at time step 
$t$, and $N$ is the horizon (possibly infinite) of the sequential decision-making problem. The maximization is taken with respect
to all possible policies $\pi$ that select an action given a state.

If the optimal Q-function is known, an agent can simply select the action that maximizes $Q^*(s,a)$ in each state in order to behave optimally and
obtain the optimal policy $\pi^*(s) = \arg \max_a Q^*(s,a)$. 

Bellman's optimality equation  \cite{Bellman:1957}
defines the optimal Q-function using a recursive equation that
relates the current Q-value to the immediate reward and 
the optimal Q-function in all possible next states:

\begin{equation}
Q^{*}(s,a)= R(s,a) + \gamma \sum_{s'} T(s,a,s') \max_b Q^*(s', b) \label{eq:Qvalues-1}
\end{equation}

Value iteration is a dynamic programming method that randomly initializes the
Q-function and then uses 
Bellman's optimality equation to compute the optimal policy
through an iterative process:

\begin{equation}
Q_{i+1}(s,a)\leftarrow R(s,a) + \gamma \sum_{s'} T(s,a,s') \max_b Q_i(s', b) 
\end{equation}

When this update is done enough times for all state-action values, it
has been shown that  $Q_i$ will converge to $Q^{*}$.

A problem with such dynamic programming methods is that usually the
transition function is not known beforehand. Furthermore, for very
large state spaces the algorithm would take an infeasible amount of
time. Therefore, reinforcement learning algorithms are more efficient
as they allow an agent to interact with an environment and learn to
optimize the policy while focusing on the most promising regions of
the state-action space.

\subsection{Q-learning}

Q-learning \cite{WatkinsQlearning1992} is one of the best-known
RL algorithms. The algorithm is based on keeping track of state-action values (Q-values)
that estimate the discounted future reward intake given a state and selected action. In the beginning the Q-function is randomly initialized and
the agent selects actions based on its Q-function and the current state. 
After each selected action, the agent obtains an experience tuple $(s, a, s', r)$ where $s'$ is the next state and $r$ the obtained reward. 
Q-learning then adjust the Q-function based on the experience in the
following way:

\begin{equation}
Q(s,a)= Q(s,a) + \alpha (r + \gamma \max_b Q(s',b) - Q(s,a)) \label{eq:Q-learning01}
\end{equation}
Where $\alpha$ is the learning rate.

During the learning phase, the agent improves its estimations and after
an infinite amount of experiences of all state-action pairs the
Q-values will converge to the optimal Q-values $Q^*(s,a)$ \cite{WatkinsQlearning1992}.
In order to learn the optimal policy, the agent has to make a trade-off
between exploitation and exploration. When the agent exploits its
currently learned Q-function, the agent selects the action with the
highest value. Although exploiting the Q-function is expected to
lead to the highest sum of future rewards, the agent also has to
perform exploration actions by selecting actions which do not have
the highest Q-value. Otherwise, it will not experience all possible
state-action values and most probably not learn the optimal policy.
In many applications, the $\epsilon$-greedy exploration policy is used
that selects a random action with probability $\epsilon$ and
the action with the highest Q-value otherwise.

\subsection{Q-learning with neural networks}

Before we explained the use of lookup tables to store the Q-values.
For very large or continuous state spaces this is not possible.
In such problems, RL is often combined with neural networks
that estimate the Q-value of each possible action given the
state representation as input. Combining an RL algorithm with
a neural network has been used
many times in the past, such as for learning to play backgammon at
human expert level \cite{Tesauro:94} in 1994. 

Q-learning can be combined with
a multi-layer perceptron (MLP) by constructing an MLP with
the same number of outputs as there are actions. 
Each output represents the Q-value $Q(s,a)$ for a particular action given the current state as input. These inputs can then be continuous and
high-dimensional and the MLP will learn a Q-function that generalizes
over the entire input space. For training the MLP on an
experience $(s, a, s', r)$,   the target-value for the
output of the selected action $a$ is defined as:

\begin{equation}
    TQ(s, a) = r + \gamma \max_b Q(s', b, \theta)
\end{equation}
Where $\theta$ denotes all adjustable weights of the MLP.
The MLP is trained using backpropagation to minimize the error 
$(TQ(s, a) - Q(s, a, \theta))^2$.
Note that the MLP is randomly initialized and is trained on its own output (bootstrapping). This often causes long training times and can
also lead to instabilities.

Mnih et al. made the combination of deep neural networks and RL
very popular by showing that such systems are very effective for
learning to play Atari games using high-dimensional input spaces (the pixels) \cite{MnihHumanlevelcontroldeep2015}.
To deal with unstable learning, they used experience replay \cite{Lin:1993},
in which a memory buffer is used to store previous experiences and
each time the agent learns on a small random subset of experiences.
Furthermore, they introduced a target network, which is a 
periodically copied version of
the Q-network. It was shown that training the Q-network not on its
own output, but on the values provided by the target network, made
the system more stable and efficient.

\section{Enhancing Q-Learning with Reward Response Filters}
There are many decision-making problems in which an action
does not have an immediate effect on the environment, but the
effect and therefore also the goodness of the action is only shown after
some delay. Take for example an archer that aims a bow and shoots an
arrow. 
If the arrow has to travel a long distance, there would be no
immediately observable reward related to the actions of the archer.
In structural control, a similar problem occurs. When a device
is controlled to produce counter forces against vibrations in a
structure caused by an earthquake, then the effects of the control
signals or actions incur a delay. This makes it difficult to
learn the correct Q-function and optimize the policy.

This paper introduces an enhanced Q-learning method
for solving sequential decision-making problems in which the maximal
effects of actions occur after a specific time-period or delay.
In this method, the agent initially builds a finite
reward response filter based on the response of the environment to a single action. Afterwards, that function is used for determining
the reward for a selected action.
The proposed method initially allows the agent to make some observations
about the behavior of the environment, so that it can later better 
determine the feedback of the executed actions. As a result,
the agent should be able to learn a better Q-function and increase its performance. For example, if the agent knows that an
action $a_{t}$, taken at time step $t$, will affect the environment
at time step $t+5$, there is no reason to use the obtained rewards
in the range of $\left[t,t+4\right]$ for determining the action value
for the selected action, $a_{t}$.

\begin{figure}[h]
\begin{centering}
\includegraphics[width=1\linewidth]{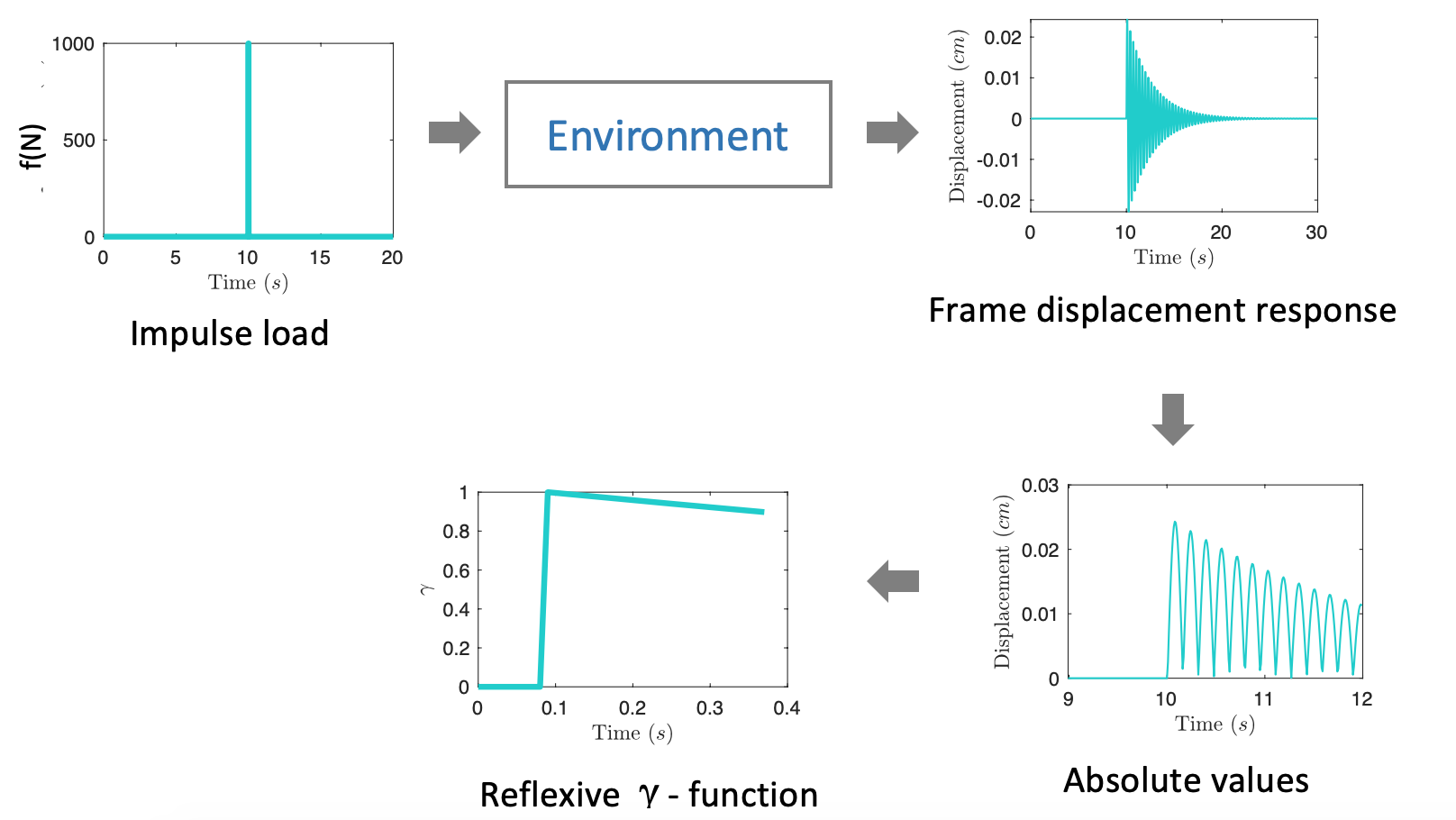}

\caption{Process of creating the reflexive $\gamma$-function. First some impulse is emitted to the environment after which the response of the environment is observed. The response is transformed to absolute
values and the start of this response is examined in detail. Finally, the finite reward response filter is determined by only keeping the peak response values. \label{fig:Process-of-creating}}
\end{centering}
\end{figure}

In the developed method, the agent takes
a single action at the beginning of the training and observes the response of the environment. Based on the observed effects, 
the agent builds a finite reward response filter, also called
reflexive $\gamma$-function,
that reflects the influence of the actions on the environment as
shown in Figure \ref{fig:Process-of-creating}. The computed
function is a normalized response of the environment
to the single taken action. As an example, in the structural control problem, the agent (controller system) applies a unit force to the roof
of the building and observes the displacement responses as a
function of time. Then, it considers the absolute values of the peak responses to  build the reward response filter. 
After the function reaches its maximum value, it will be gradually decreased to zero. As the most important point is the time at which the action effect reaches its maximum, the gamma function is cut after
the time step in which the response of the environment is reduced by 
$p$ percent of its peak value. In the experiments we set $p = 15$ percent.
Note that the main idea can be used for other problems by
constructing different ways of creating the reward response filter.

During the learning phase,
the agent computes the target Q-values $TQ(s, a)$ by multiplying the reflexive $\gamma$-function  with the immediate rewards during the time interval $[t_{0},t_{0}+\Delta t_{\gamma}]$ and adding the maximum Q-value of the successor state $s' = s_{t_{m})}$ as follows:

\begin{equation}
TQ(s, a)= \stackrel[j=0]{n}{\sum}(\gamma_{j}r_{j})+\gamma_{n+1}\,\underset{a}{max}\,Q(s',a')\label{eq:Q-learning01-1}
\end{equation}

\noindent  where $\gamma_j$ values are obtained from the reflexive $\gamma$-function, and:

$r_{j}$: immediate reward value  at time $t_{0}$

$n$: number of the time steps in reflexive $\gamma$-function.

$t_{0}:$ time when agent takes action $a$

$t_{m}:$ time step after $t_{0}+\Delta t_{\gamma}$

$\Delta t_{\gamma}$ : time duration of reflexive $\gamma$-function

$s'$: successor state at time $t_{0}+\Delta t_{\gamma}$

Note that our method is related to multi-step Q-learning \cite{RichardS.SuttonReinforcementLearningintroduction1998} in which
the Q-function is updated based on the rewards obtained in multiple
steps. The differences are: 1) In the proposed method, the time
duration of the finite reward response filter is determined
automatically at the start of the experiment. 2) The proposed
method does not use exponential discounting of future rewards, but
determines the discount factors based on the observed effects of
actions on the environment. 3) It is possible that the immediate rewards after taking an action have zero gamma values and therefore are excluded by the algorithm in updating the Q-value. 

\section{Case Study}
As a case study, we train an agent as a structural controller
to reduce the vibrations of a single-story building, subjected to   earthquake
excitations. The structural system of the building comprises a moment
frame which is modeled as a single degree of freedom (SDOF) system.
The mathematical model of the structure (the environment) is developed
in Simulink software in which an analysis module determines the structural responses. The mass, stiffness, and damping of the structure are the constant values of the model during the simulations. The inputs to the analysis module include the earthquake acceleration record as an external excitation and the control forces which are the transformed values of the control signals, generated by the neural network. The outputs of the analysis module
are the displacement, velocity and the acceleration responses of the frame.

Action-effect delays are considered in the model using a \emph {delay-function} which applies
input excitation to the frame after a constant delay. 
Based on the value of the delay between the occurrence of an  external excitation and when it affects the structure, three simulation scenarios including no delay, medium delay (5 seconds) and a long delay (10
seconds) are considered and  the performance of the enhanced method is compared to the original method.

\subsection{Agent}
The intelligent controller consists of a neural network called \emph{Q-net} which receives as inputs the current state, including the structural responses and the external excitation from the analytical model, and generates
 the Q-values. The action corresponding to the maximum Q-value will then be sent to the actuator module which transforms the control signals to the force values and passes that to the analysis module. In addition to Q-net, a secondary stabilizer (target)
net is also developed to improve the performance of the learning
module, as was proposed by Volodymyr Mnih et al. \cite{MnihHumanlevelcontroldeep2015}
in the experience replay learning method. During the learning phase the Q-net is trained to improve its performance by learning the Q-values. The training algorithm and the hyper-parameters are presented in Table \ref{tab:Neural-network's-hyper}.

\begin{table}[H]
\begin{centering}
\begin{tabular}{>{\centering}m{0.15\linewidth}>{\centering}m{0.15\linewidth}>{\centering}p{0.25\linewidth}>{\centering}p{0.25\linewidth}}
\hline 
\multicolumn{2}{c}{hidden layers} & \multirow{2}{0.25\linewidth}{\centering{}training algorithm} & \multirow{2}{0.25\linewidth}{\centering{}learning rate}\tabularnewline
\cline{1-2} \cline{2-2} 
number & size &  & \tabularnewline
\hline 
2 & 40 & back propagation & 0.99\tabularnewline
\hline 
\end{tabular}
\par\end{centering}
\centering{}\caption{Neural network's training parameters\label{tab:Neural-network's-hyper}}
\end{table}

\subsection{Environment}

\begin{figure}[H]
\begin{centering}
\includegraphics[width=0.8\linewidth]{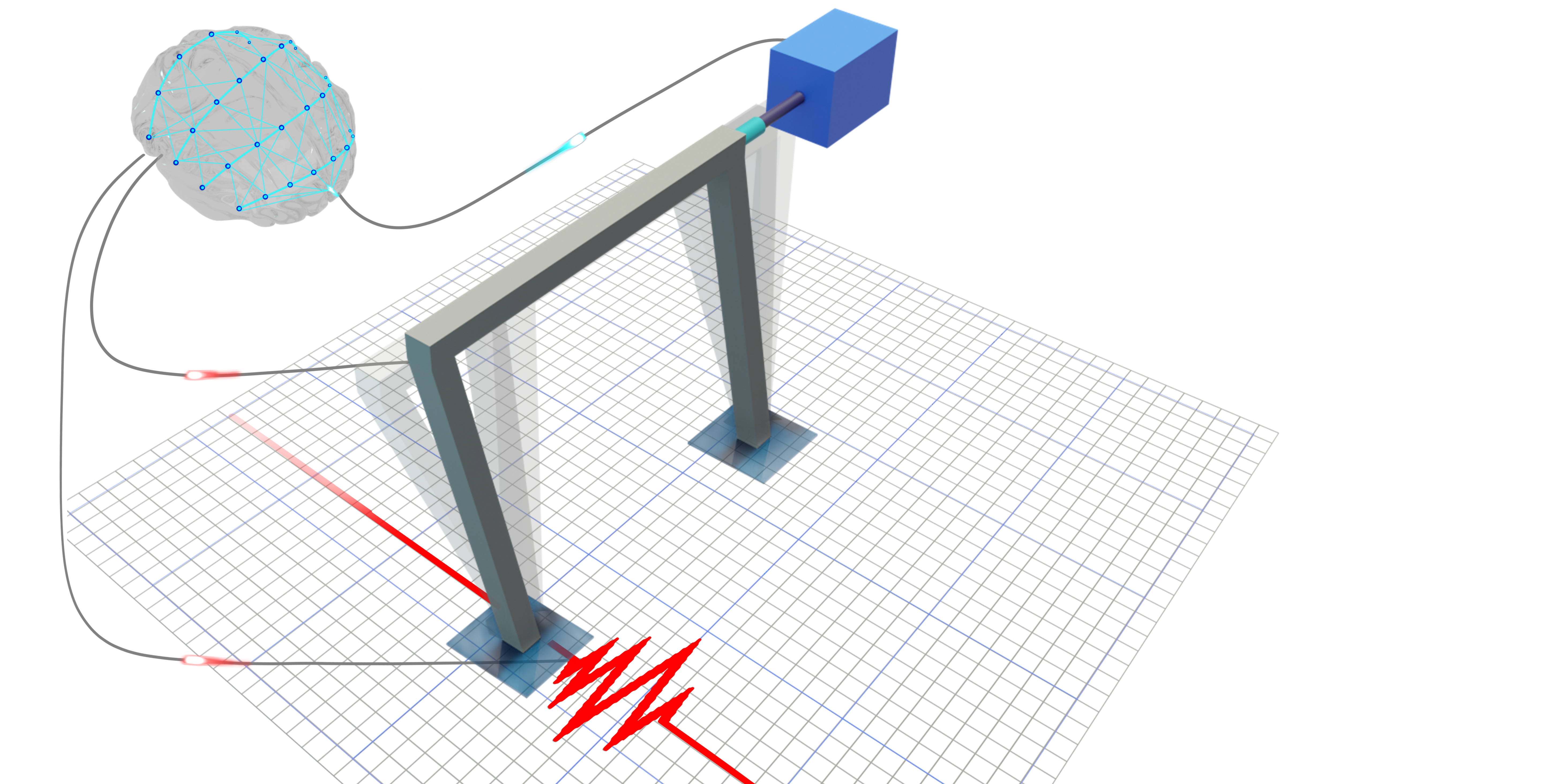}
\par\end{centering}
\caption{Schematic presentation of the structural control problem \label{fig:The-simulation-environment}}
\end{figure}

The concept of the intelligent control of the frame is schematically shown in Figure \ref{fig:The-simulation-environment}. As it is shown, a moment frame is subjected to the earthquake excitation and the control system transforms the control signals from the neural network to the horizontal forces on the frame using an actuator.  
The mass of the frame is $2000$ kg, which is considered as a condense mass at the roof level, the stiffness
is $7.9\times10^{6}$ {[}N/s{]}, and the damping is $250\times10^{3}$
N.S/m. The natural frequency $\omega$, and the period
$T$ of the system are as follows:

\begin{equation}
\omega=\sqrt{\frac{k}{m}}=\sqrt{\frac{7.9e6}{2000}}=62.84\:\frac{1}{s}
\end{equation}

\begin{equation}
T=\frac{2\pi}{\omega}=0.1\:s
\end{equation}

\subsection{Environment dynamics}

The Equation of the motion for the moment frame under the earthquake excitation
and the control forces is as follow:

\begin{equation}
m\ddot{\mathbf{\mathrm{\mathrm{u}}}}+c\mathrm{\dot{u}}+k\mathrm{u}=-m\ddot{x}_{g}+f\label{eq:State Space-01}
\end{equation}

\noindent in which $m,\:c,\,k$ are the mass, damping and the stiffness
matrices, $\ddot{x}_{g}$ is the ground acceleration,
 and $f$ is the control force. $u$, $\dot{u},$ and $\ddot{u}$ are
displacement, velocity, and acceleration vectors respectively.

\noindent By defining the state vector $x$ as:

\begin{equation}
x=\left\{ \mathrm{u},\,\mathrm{\dot{u}}\right\} ^{T}\label{State 01}
\end{equation}

\noindent The state-space representation of the system would be:

\begin{equation}
\dot{x}=Ax+Ff+G\ddot{x}_{g}\label{eq:SS-03}
\end{equation}

\begin{equation}
y_{m}=C_{m}x
\end{equation}

\noindent Considering $v=\left\{ \ddot{x}_{g},\,f\right\} $, Equation \ref{eq:SS-03}
can be written as:

\begin{equation}
\dot{x}=Ax+Bv
\end{equation}

\noindent in which:

\begin{eqnarray*}
A & = & \left[\begin{array}{cc}
0 & 1\\
-\frac{k}{m} & -\frac{c}{m}
\end{array}\right]=\left[\begin{array}{cc}
0 & 1\\
-3947.8 & -125.66
\end{array}\right]\\
B & = & \left[\begin{array}{cc}
0 & 0\\
-1 & \frac{1}{m}
\end{array}\right]=\left[\begin{array}{cc}
0 & 0\\
-1 & 5\times10^{-4}
\end{array}\right]\\
C_{m} & = & \left[\begin{array}{cc}
1 & 0\\
0 & 1
\end{array}\right]
\end{eqnarray*}

\subsection{Earthquake excitation}

\begin{figure}[H]
\begin{centering}
\includegraphics[width=0.5\linewidth]{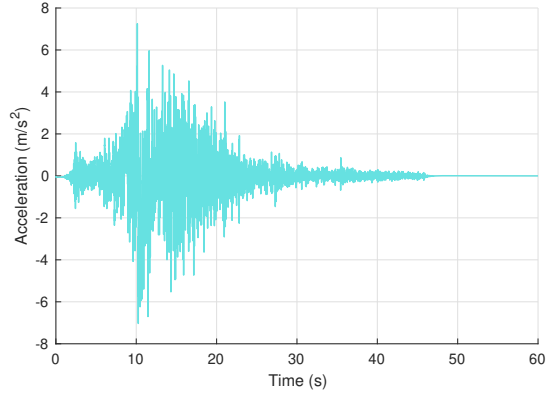}
\par\end{centering}
\caption{Landers earthquake record, obtained from Pacific Earthquake Engineering
Research (PEER) Center \cite{PEERGroundMotion} \label{fig:Landers-earthquake-record}}
\end{figure}

In order to train the intelligent controller, the acceleration record of Landers earthquake is considered which is obtained from the NGA strong motion database \cite{PEERGroundMotion}
(See Figure \ref{fig:Landers-earthquake-record}).

\subsection{State\label{subsec:State-1}}

The state shall include all the required data for determining the actions to be an Markov state. In this regard, for the structural control problem, the ground accelerations as well as the structural
responses including acceleration, velocity, and displacement responses
are included in the each state $S_{t}$. Based on the author's experience, including the displacements in the
last three time steps helps the agent to understand the direction
of the motion which is important specially when reaching the maximums
in the oscillations.

\begin{equation}
\left\{ u_{t},u_{t-1},u_{t-2},v_{t},a_{t},\ddot{u}_{g,t}\right\} \epsilon\,S_{t}\label{eq:state definition-1}
\end{equation}

\noindent where:

$u_{t}$ : displacement in time t

$v_{t}$ : velocity in time t

$a_{t}$ : acceleration in time t

$\ddot{u}_{g,t}$: ground acceleration in time t

\subsection{Reward Function}
In reinforcement learning, the reward function determines the goodness of the 
taken action. In this
regard, a multi-objective reward function is defined including four
partial rewards:

\subsubsection*{Displacement response}

The first partial reward function reflects the performance of the
controller in terms of reducing the displacement responses:

\[
R_{1,t}=1-\frac{\mid u_{t}\mid}{u_{max}}
\]

\noindent  in which, $u_{t}$ is the displacement value of the frame
at time\emph{ $t$ }and $u_{max}$ is the maximum uncontrolled
displacement response.

\subsubsection*{ Velocity response}

This partial reward evaluates the velocity response of the frame:

\[
R_{2,t}=1-\frac{\mid v_{t}\mid}{v_{max}}
\]

\noindent  in which $v_{t}$ is the velocity response of the frame
at time\emph{ $t$} and $v_{max}$ is the maximum uncontrolled
velocity response.

\subsubsection*{ Acceleration response}

The performance of the controller in term of reducing the acceleration
responses of the frame is evaluated by $R_{3,t}$:

\[
R_{3,t}=1-\frac{\mid a_{t}\mid}{a_{max}}
\]

\noindent  in which $a_{t}$ is the acceleration response of the
frame at time\emph{ $t$} and $a_{max}$ is the maximum uncontrolled
acceleration response.

\subsubsection*{Actuator force}
The goal of the fourth partial reward is to evaluate the required energy
by applying a penalty value equal to $0.005$ to the actuator force
in each time-step:

$R_{4,t}=f_{t}\times P_{a}$

\noindent in which:

$f_{t}=$ Actuation force at time $t$ ($N$)

$P_{a}=$Penalty value for unit actuator force ($=0.005$)

By combining the four partial rewards, the reward value $R$, at 
time $t$ will be calculated:

\[
R_{t}=R_{1,t}+R_{2,t}+R_{3,t}+R_{4,t}
\]

\section{Training}

Following the experience-replay method, in each training episode, many experiences were recorded in the experience-replay memory buffer. The algorithm then randomly selected 100 of the records from the experience replay buffer and experiences were selected by the key state selector function as proposed 
in  \cite{Rahmaniframeworkbrainlearningbased2019} to train the Q-net.  The target network was updated every 50 training episodes. The
learning parameters have been tuned during preliminary experiments
and are shown in Table \ref{tab:Utilized-Learning-parameters}.

For balancing the exploration and exploitation during the learning
phase, the agent followed the $\epsilon$-greedy policy which means that in
each state, it took the action with the maximum expected return, but
occasionally it took random actions with a probability of $\epsilon$. In this research, a decreasing $\epsilon$-value technique is utilized so that its value gradually reduces from  1 in the beginning to a minimum value of 0.1.

\begin{table}[tbh]
\caption{Utilized Learning parameters\label{tab:Utilized-Learning-parameters}}

\centering{}%
\begin{tabular}{>{\centering}p{0.15\linewidth}>{\centering}p{0.15\linewidth}>{\centering}p{0.15\linewidth}>{\centering}p{0.15\linewidth}>{\centering}p{0.15\linewidth}}
Number of episodes & Size of experience buffer & Number of states per episode & Sensor sampling rate $(Hz)$ & Mini-batch size\tabularnewline
\hline 
1000 & 60000 & 6000 & 100 & 50\tabularnewline
\end{tabular}
\end{table}

\section{Results}

\begin{figure*}[hp]
\begin{centering}
\textsf{}\subfloat[Original method - 0 s delay]{\begin{centering}
\includegraphics[width=0.40\linewidth]{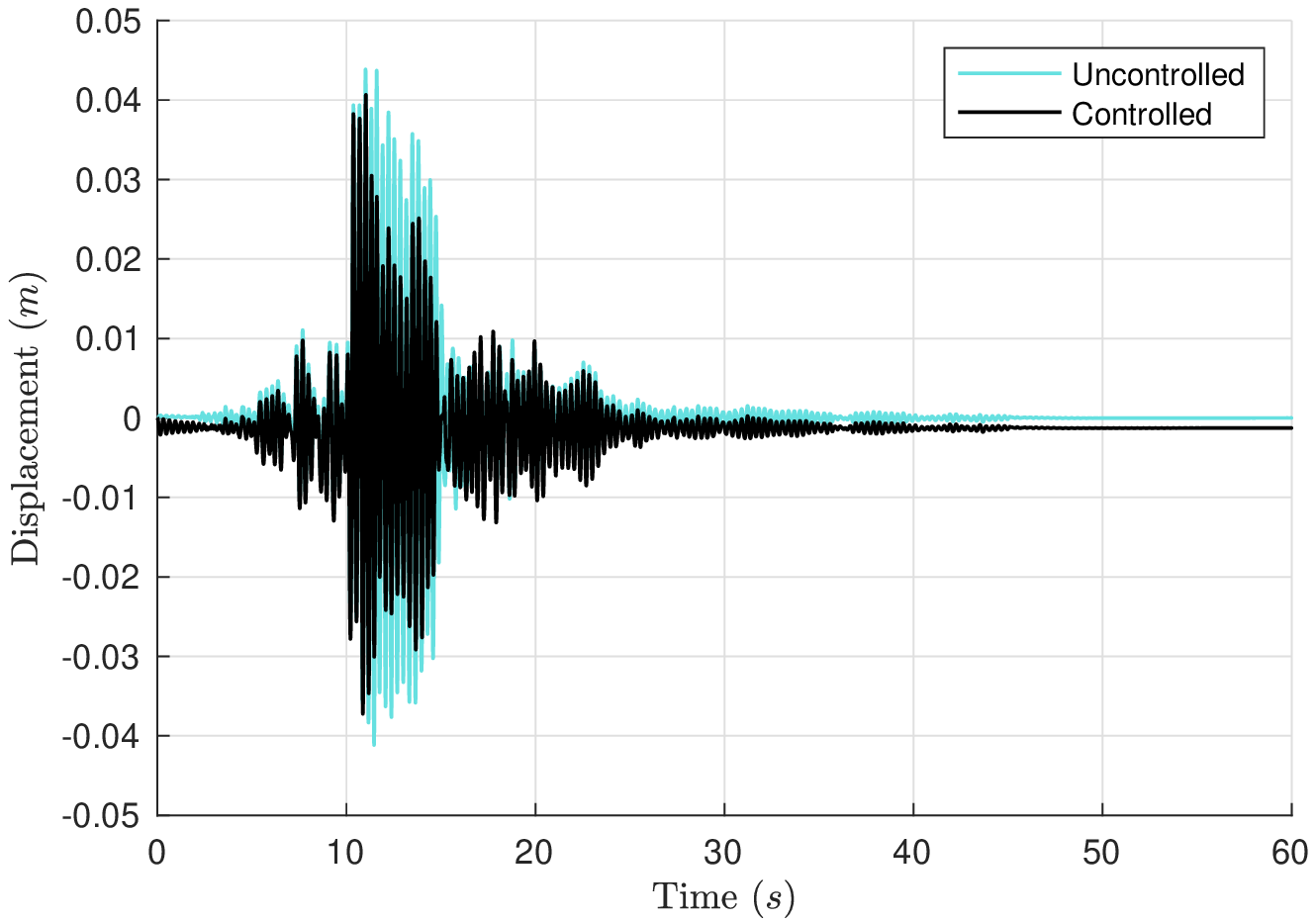}
\par\end{centering}
\textsf{}}\textsf{}\subfloat[Enhanced method - 0 s delay]{\begin{centering}
\includegraphics[width=0.45\linewidth]{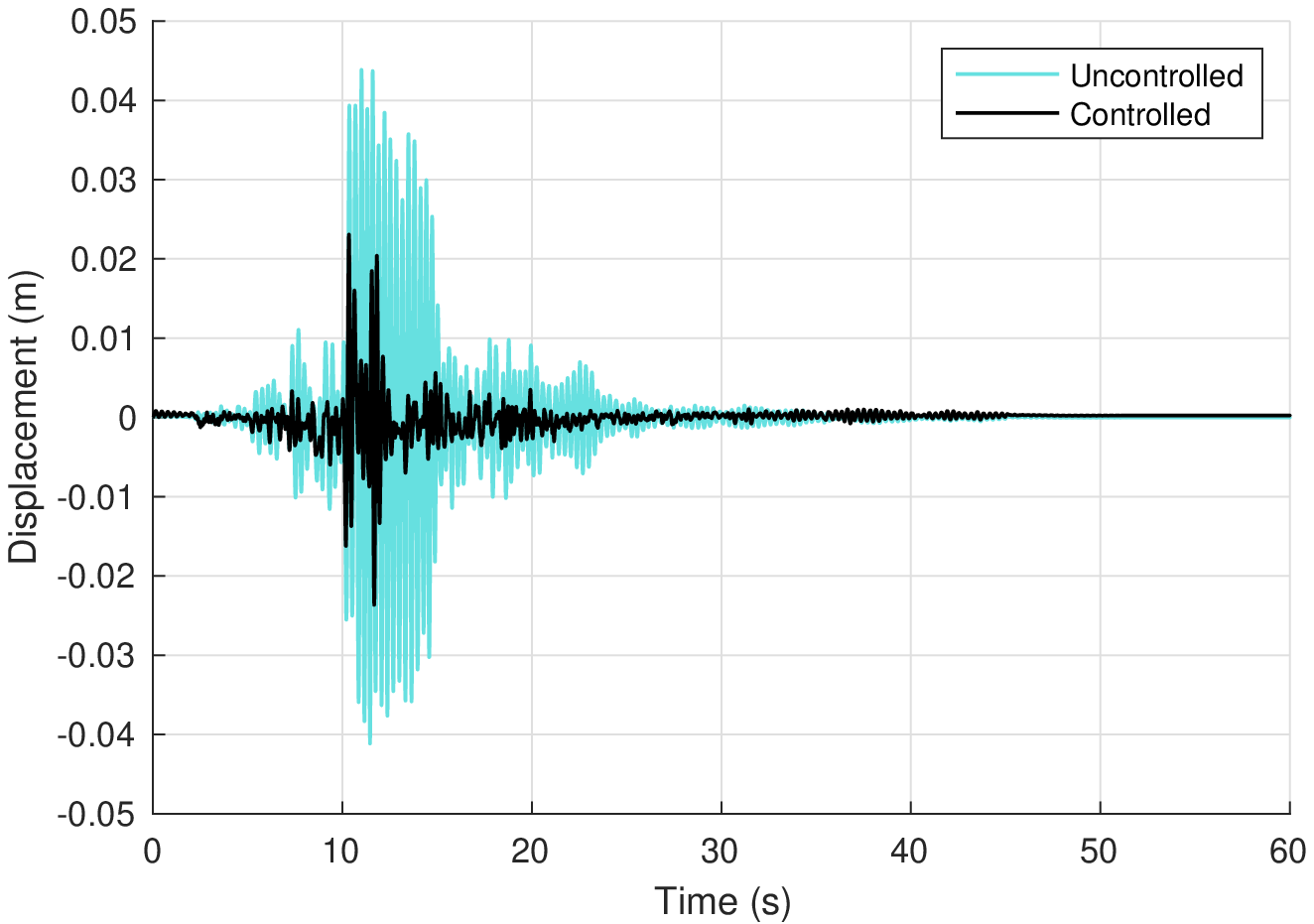}
\par\end{centering}
\textsf{}}
\par\end{centering}
\begin{centering}
\textsf{}\subfloat[Original method - 5 s delay]{\begin{centering}
\includegraphics[width=0.45\linewidth]{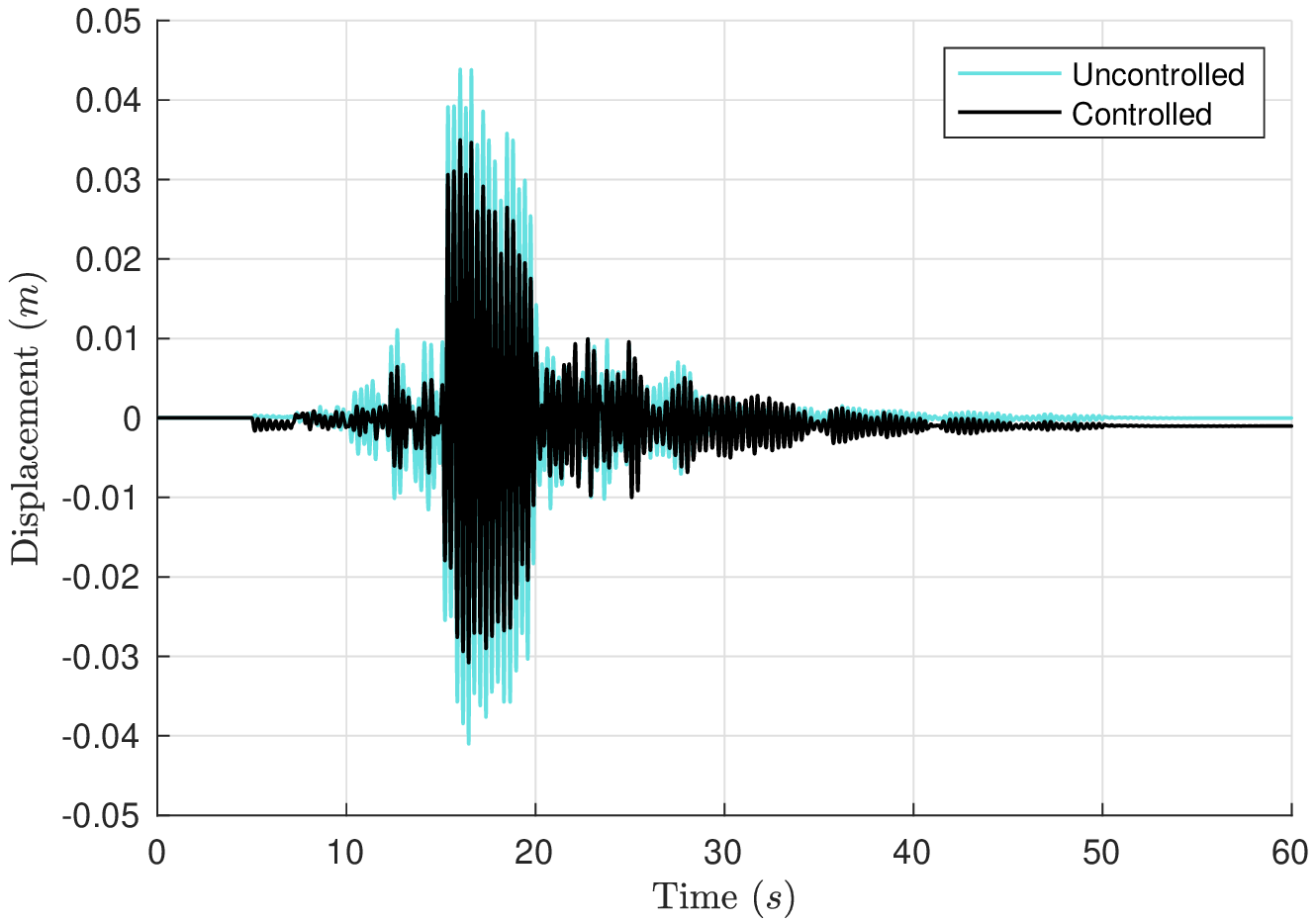}
\par\end{centering}
\textsf{}}\textsf{}\subfloat[Enhanced method - 5 s delay]{\begin{centering}
\includegraphics[width=0.45\linewidth]{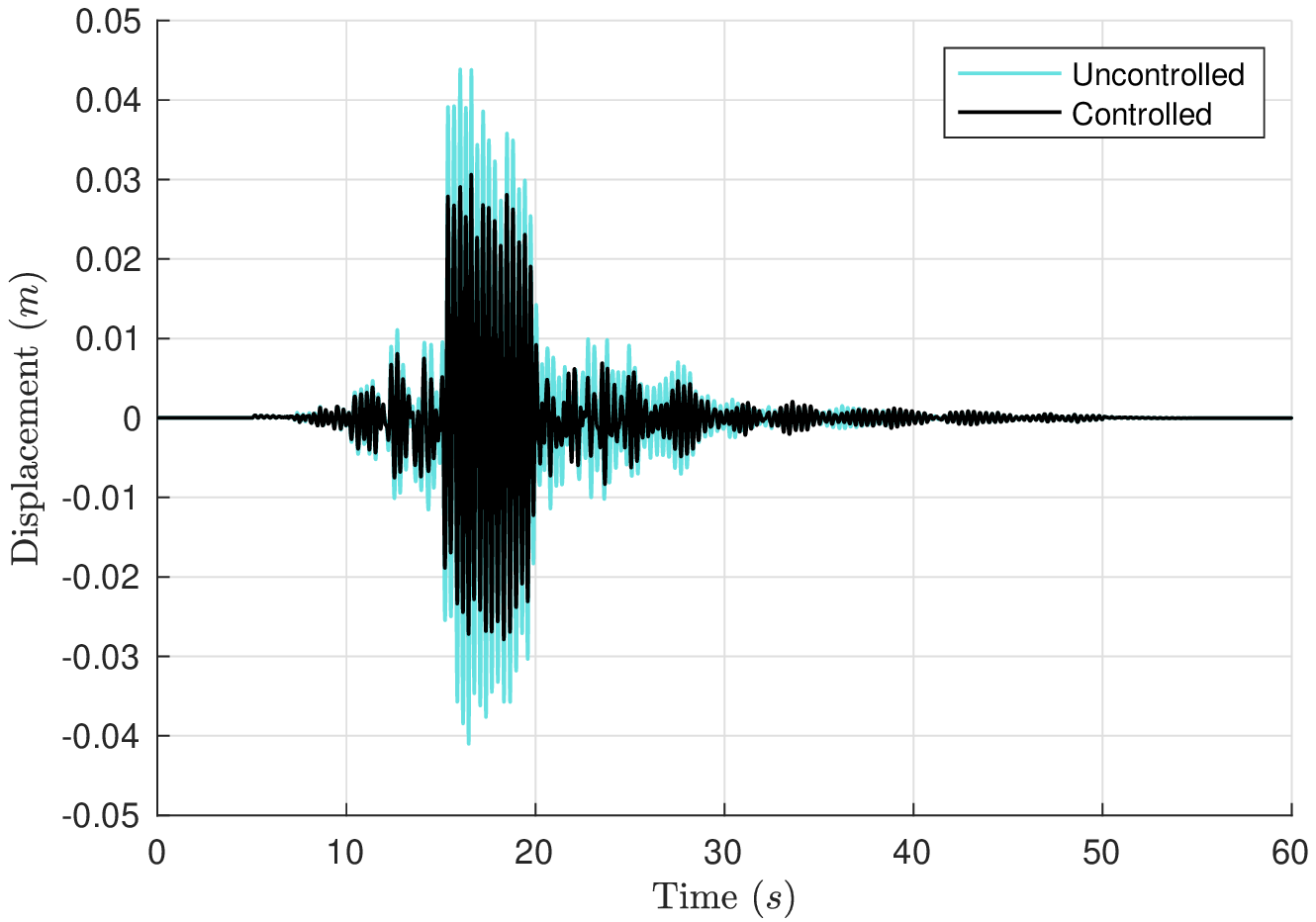}
\par\end{centering}
\textsf{}}
\par\end{centering}
\begin{centering}
\textsf{}\subfloat[Original method - 10 s delay]{\begin{centering}
\includegraphics[width=0.45\linewidth]{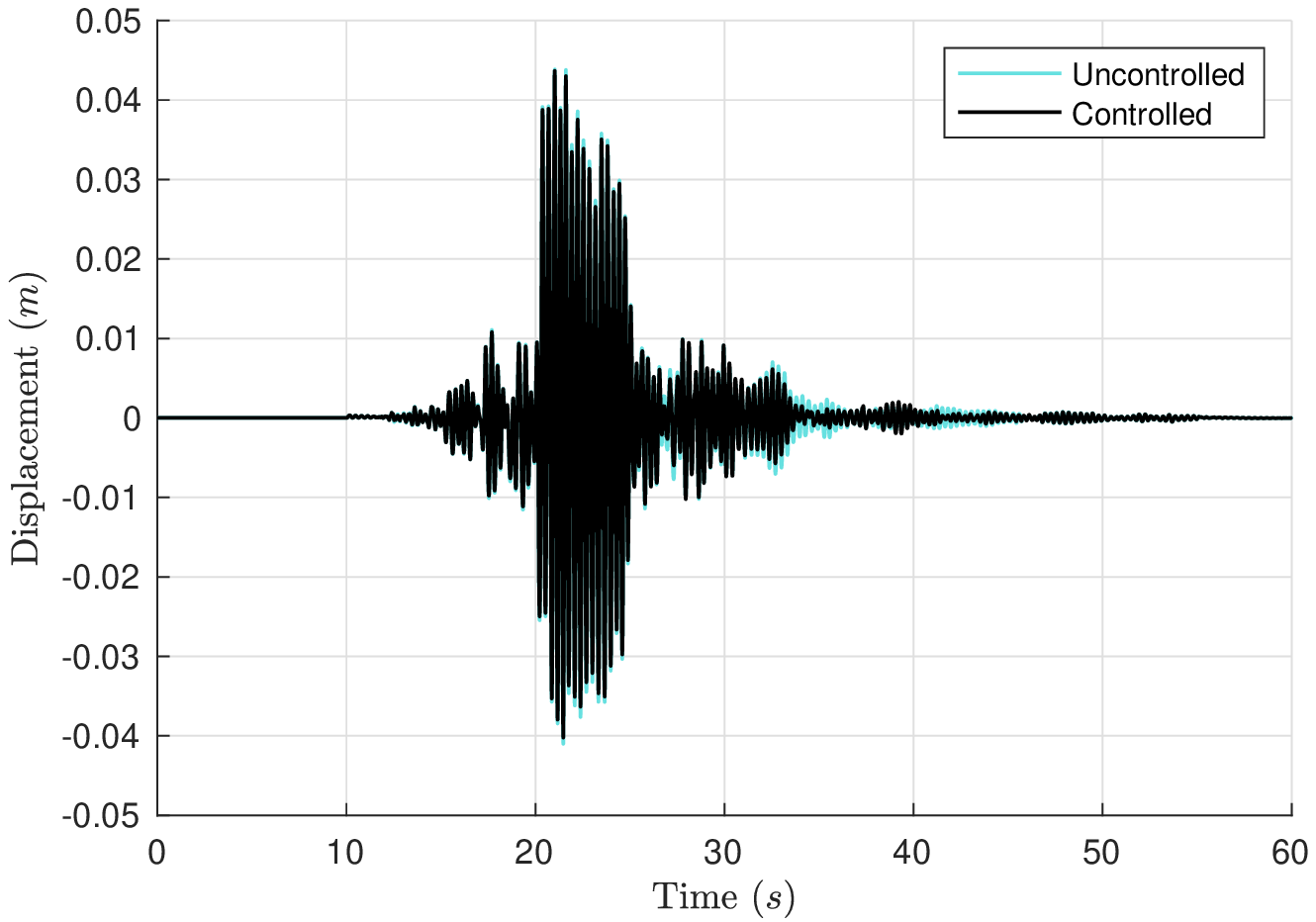}
\par\end{centering}
\textsf{}}\textsf{}\subfloat[Enhanced method - 10 s delay]{\begin{centering}
\includegraphics[width=0.45\linewidth]{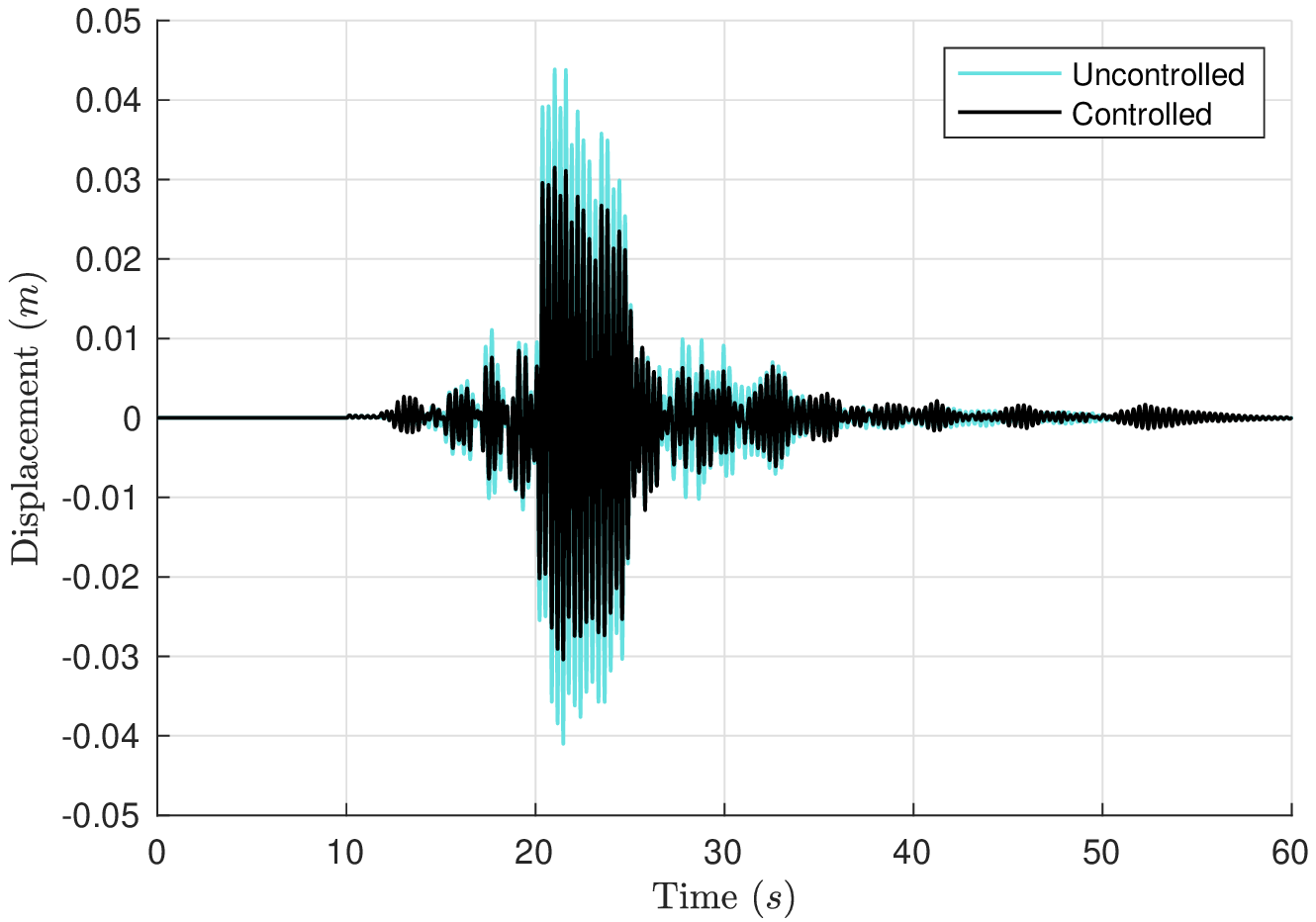}
\par\end{centering}
\textsf{}}
\par\end{centering}
\caption{Uncontrolled/controlled roof displacement response of the frame when
the controller was trained using the original method and enhanced method.\label{fig:Uncontrolled/controlled-roof-dis}}
\end{figure*}

During the learning phase, the controller was trained for three scenarios including zero delay ($\Delta t_{d} = 0\: s$),  medium delay ($\Delta t_{d} = 5 \: s$), and long delay ($\Delta t_{d} = 10 \: s$). In each scenario, the original and the enhanced methods were utilized and the maximum-achieved performance of the controller was recorded as presented in Tables \ref{tab:Comparing-the-performance-Improved-1-1}
to \ref{tab:Comparing-the-performance-Improved}. The corresponding uncontrolled
and controlled displacement responses of the frame are demonstrated
in Figure \ref{fig:Uncontrolled/controlled-roof-dis}. The results
show that the enhanced method has significantly improved the performance
 in all scenarios. Even without action-effect delays,
the performance of the enhanced method in terms of reducing the displacement
responses is 46.1\% which is much better than the original method 
that minimizes the displacement with 7.1\%.

\begin{table}[htbp]
\caption{Seismic responses of the fame in case of zero delay. (Dis. = Displacement
$(cm),$ Vel. = Velocity $(m/s)$, Acc. = Acceleration $(m/s^{2})$).\label{tab:Comparing-the-performance-Improved-1-1}}

\begin{centering}
\setlength\tabcolsep{1.5pt}
\par\end{centering}
\centering{}%
\begin{tabular}{>{\raggedright}m{0.1\linewidth}>{\centering}p{0.14\linewidth}>{\raggedright}p{0.12\linewidth}>{\raggedright}p{0.12\linewidth}>{\raggedright}p{0.12\linewidth}}
Learning method &  & \multirow{1}{0.12\linewidth}{\centering{}Uncontrol.} & \multirow{1}{0.12\linewidth}{\centering{}Controlled} & \multirow{1}{0.12\linewidth}{\centering{}Improvement}\tabularnewline
\midrule
\multirow{3}{0.1\linewidth}{Original method} & Peak Dis. & 4.39 & 4.06 & 7.1\%\tabularnewline
 & Peak Vel. & 0.91 & 0.83 & 8.7\%\tabularnewline
 & Peak Acc. & 22.97 & 16.84 & 26.7\%\tabularnewline
\midrule
\multirow{3}{0.1\linewidth}{Enhanced method} & Peak Dis. & 4.39 & 2.36 & 46.1\%\tabularnewline
 & Peak Vel. & 0.91 & 0.54 & 41.0\%\tabularnewline
 & Peak Acc. & 22.97 & 14.28 & 37.8\%\tabularnewline
\bottomrule
\end{tabular}
\end{table}

\begin{table}[htbp]
\caption{Seismic responses of the fame when the delay is 5 seconds (Dis. =
Displacement $(cm),$ Vel. = Velocity $(m/s)$, Acc. = Acceleration
$(m/s^{2})$).\label{tab:Comparing-the-performance-Improved-1}}

\begin{centering}
\setlength\tabcolsep{1.5pt}
\par\end{centering}
\centering{}%
\begin{tabular}{>{\raggedright}m{0.1\linewidth}>{\centering}p{0.14\linewidth}>{\raggedright}p{0.12\linewidth}>{\raggedright}p{0.12\linewidth}>{\raggedright}p{0.12\linewidth}}
Learning method &  & \multirow{1}{0.12\linewidth}{\centering{}Uncontrol.} & \multirow{1}{0.12\linewidth}{\centering{}Controlled} & \multirow{1}{0.12\linewidth}{\centering{}Improvement}\tabularnewline
\midrule
\multirow{3}{0.1\linewidth}{Original method} & Peak Dis. & 4.39 & 4.24 & 3.4\%\tabularnewline
 & Peak Vel. & 0.91 & 0.85 & 6.5\%\tabularnewline
 & Peak Acc. & 22.93 & 20.0 & 12.8\%\tabularnewline
\midrule
\multirow{3}{0.1\linewidth}{Enhanced method} & Peak Dis. & 4.39 & 3.06 & 30.2\%\tabularnewline
 & Peak Vel. & 0.91 & 0.62 & 32.2\%\tabularnewline
 & Peak Acc. & 22.93 & 16.38 & 28.5\%\tabularnewline
\bottomrule
\end{tabular}
\end{table}

\begin{table}[htbp]
\caption{Seismic responses of the fame when the delay is 10 seconds (Dis. =
Displacement $(cm),$ Vel. = Velocity $(m/s)$, Acc. = Acceleration
$(m/s^{2})$).\label{tab:Comparing-the-performance-Improved}}

\begin{centering}
\setlength\tabcolsep{1.5pt}
\par\end{centering}
\centering{}%
\begin{tabular}{>{\raggedright}m{0.1\linewidth}>{\centering}p{0.14\linewidth}>{\raggedright}p{0.12\linewidth}>{\raggedright}p{0.12\linewidth}>{\raggedright}p{0.12\linewidth}}
Learning method &  & \multirow{1}{0.12\linewidth}{\centering{}Uncontrol.} & \multirow{1}{0.12\linewidth}{\centering{}Controlled} & \multirow{1}{0.12\linewidth}{\centering{}Improvement}\tabularnewline
\midrule
\multirow{3}{0.1\linewidth}{Original method} & Peak Dis. & 4.39 & 4.37 & 0.41\%\tabularnewline
 & Peak Vel. & 0.91 & 0.89 & 1.81\%\tabularnewline
 & Peak Acc. & 22.93 & 22.31 & 2.70\%\tabularnewline
\midrule
\multirow{3}{0.1\linewidth}{Enhanced method} & Peak Dis. & 4.39 & 3.15 & 28.2\%\tabularnewline
 & Peak Vel. & 0.91 & 0.66 & 26.9\%\tabularnewline
 & Peak Acc. & 22.93 & 17.65 & 23.0\%\tabularnewline
\bottomrule
\end{tabular}
\end{table}

As it was expected, by increasing the action-effect delay from zero
to 5 seconds, the performance of both methods is reduced, but still
the enhanced method, that was utilizing the $\gamma$ - function for estimating
the Q-values, shows a better performance. For example, in case
of having 5 seconds action-effect delays, the performance of the enhanced
method in terms of reducing the displacement responses is 30.2\% which
is by far better than the value for the original method with 3.4\%.

By studying the performance of both methods under a long action-effect 
delay (10 seconds), it is understood that the enhanced method not only
significantly  improves the performance of the training algorithm but
also  shows a  higher stability in training an agent for different action-delays. For example, by increasing the delays from 5 seconds to
10 seconds, the performance of the original method in terms of improving
the displacement response is dropped from 3.4\% to 0.41\%, which implies 87\%
performance reduction, while the performance of the enhanced method is 
reduced from 30.2\% to 28.2\% which indicates only
7\% reduction in performance.

\begin{figure}[tbh]
\subfloat[Original method]{\includegraphics[width=0.45\linewidth]{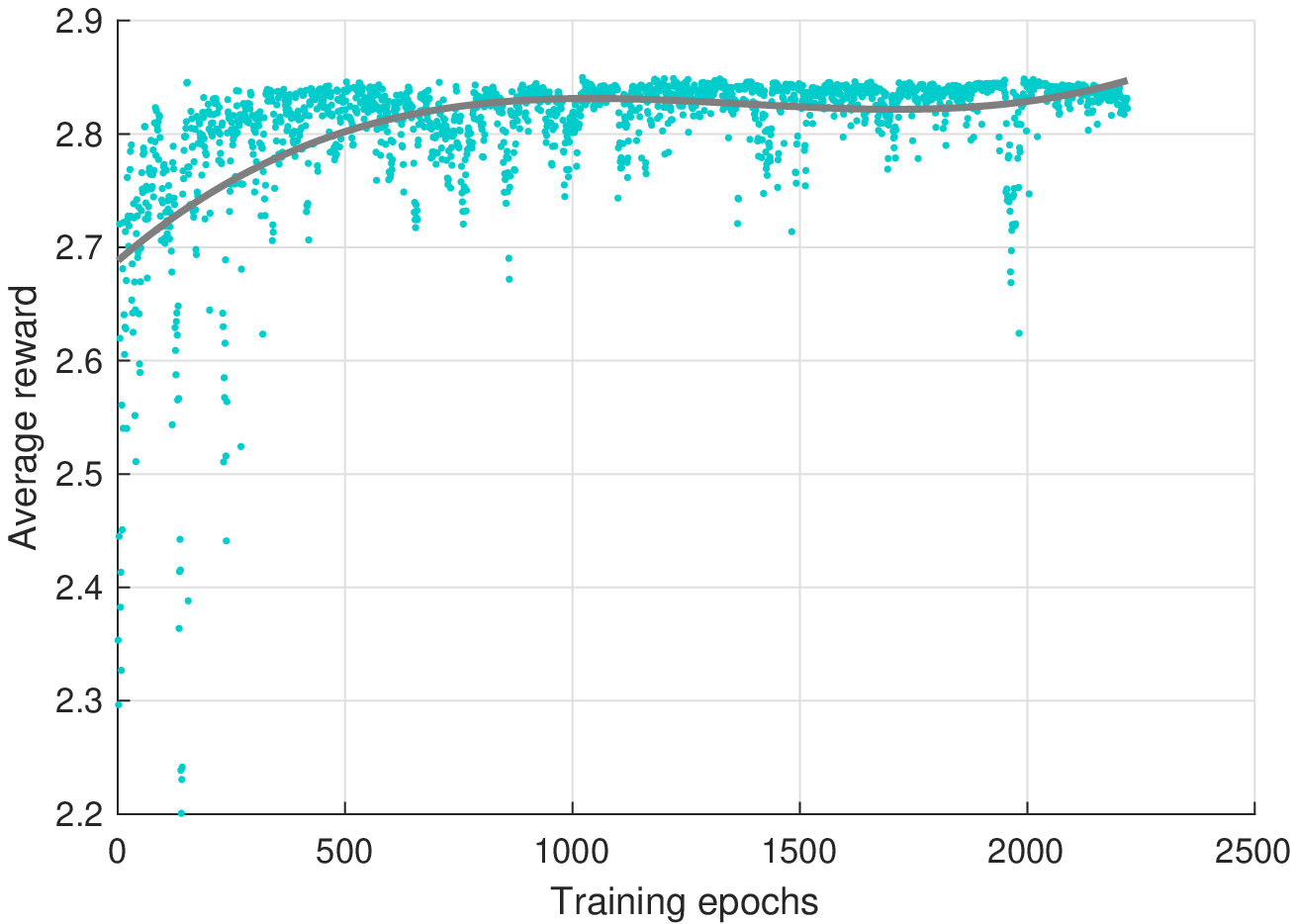}

}\subfloat[Enhanced method]{\includegraphics[width=0.45\linewidth]{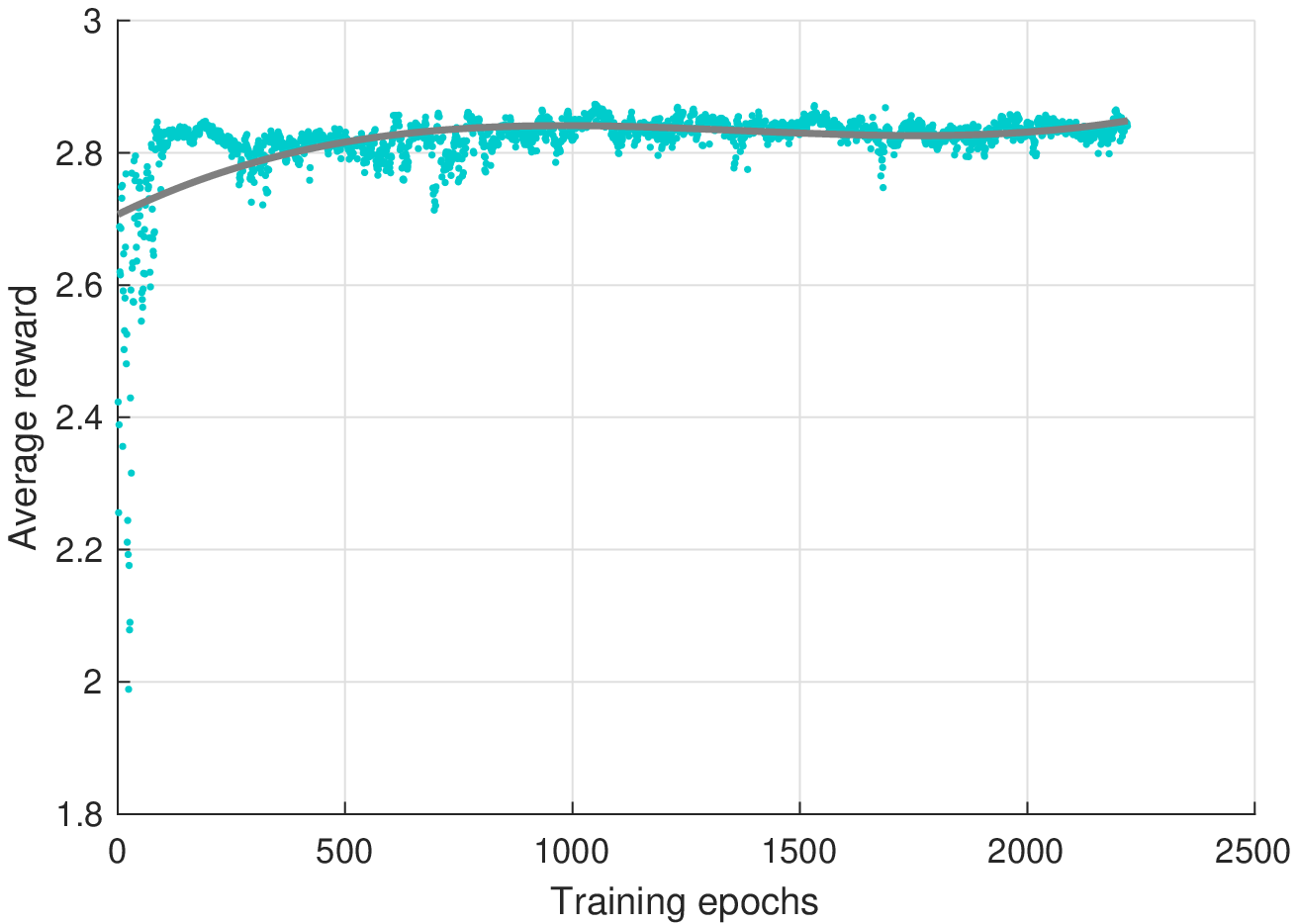}

}

\caption{Average reward during the training episodes in the experiment with 10s
delay\label{fig:Average-reward-during}}
\end{figure}

In order to better study the performance of both algorithms in
case of experiencing long action-effects delays, the average reward values
over the training episodes are presented in Figure \ref{fig:Average-reward-during}.
As it is shown, the agent achieves a much better  performance within less
training episodes when it uses the enhanced method for learning
the Q-values.

\section{Conclusions}

An enhanced Q-learning method is proposed
in which the agent creates a $\gamma$
- function  based on the environment's response to a single action at the beginning of the learning phase, and uses that to update the  Q-values in each state.
The experimental results showed that 
the enhanced algorithm significantly improved the performance and stability of the learning algorithm in dealing with action-effect delays.

Considering the obtained results, the following conclusions are drawn:
\begin{enumerate}
\item The enhanced method has significantly improved the performance of the
original method in all scenarios with different action-effect delays. Even without having any action-effect
delays.
\item  Increasing such delays results in a drop in 
the performance of the original method
while the enhanced method has shown a stable performance with a small decrease in its performance.
\item It should be mentioned that in the developed method, the $\gamma$
- function remains constant during the training phase, which implies
that the enhanced method is appropriate for environments with a static
behavioral response to the agent's actions.
\end{enumerate}

In future work, it would be interesting to study the performance
of the enhanced RL algorithm in more complex simulations using
larger structures. We would also like to use the finite response
reward filter for other problems in which there are
delays in the effects of actions.
\newpage

\bibliographystyle{plain}

\bibliography{ReflexiveGamma} 
\end{document}